# Saliency based Semi-supervised Learning for Orbiting Satellite Tracking

Peizhuo Li[#], Yunda Sun[#], and Xue Wan[*]

*Abstract*—The trajectory and boundary of an orbiting satellite are fundamental information for on-orbit repairing and manipulation by space robots. This task, however, is challenging owing to the freely and rapidly motion of on-orbiting satellites, the quickly varying background and the sudden change in illumination conditions. Traditional tracking usually relies on a single bounding box of the target object, however, more detailed information should be provided by visual tracking such as binary mask. In this paper, we proposed a SSLT (Saliency-based Semi-supervised Learning for Tracking) algorithm that provides both the bounding box and segmentation binary mask of target satellites at 12 frame per second without requirement of annotated data. Our method, SSLT, improves the segmentation performance by generating a saliency map based semi-supervised on-line learning approach within the initial bounding box estimated by tracking. Once a customized segmentation model has been trained, the bounding box and satellite trajectory will be refined using the binary segmentation result. Experiment using real on-orbit rendezvous and docking video from NASA (Nation Aeronautics and Space Administration), simulated satellite animation sequence from ESA (European Space Agency) and image sequences of 3D printed satellite model took in our laboratory demonstrate the robustness, versatility and fast speed of our method compared to state-of-the-art tracking and segmentation methods. Our dataset will be released for academic use in future.

I. Introduction

Space robots play a crucial role in current and future space exploration missions [1], as they can access to harsh environment of space and accomplish complex space tasks, such as repairing and assembling satellites, while reducing the risk of astronauts [2]. To repair or assemble a satellite in space, a fundamental step is to track and segment the target satellite using the in-situ sensors. The visual perception information of the target satellite will help our understanding the state of the satellite for decision making in further step, such as rendezvous and docking [3, 4], repairing and manipulation [5], etc. It is also important for on-board image compressing, as the irrelevant background information could be largely compressed leaving the target satellite information quickly processed by on-board and ground-based image analysis system.

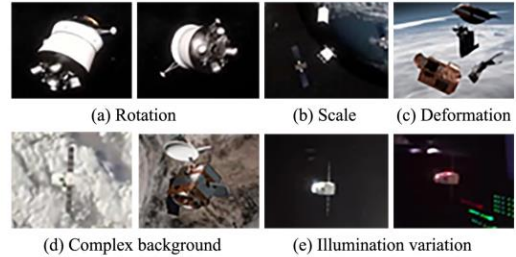

Figure 1. Examples of challenging scenes for orbiting satellite tracking and segmentation. *(figure(a) shows a satellite spin between two image frames, figure(b) shows scale change in image sequence while the target satellite is the same as figure(a) but appears much smaller in the image sequence, figure(c) shows a satellite disintegration scene which a satellite breaks into several part, figure(d) shows two scenes which background has similar texture to satellite, and figure(e) shows illumination variation between two image frames)*

Most state-of-the-art visual tracking methods can only track the center point of the object [6-11], however, the bounding box alone could hardly indicate complex motion, such as spin and rotation, and deformation of the satellite. In this paper, we narrow the gap between traditional tracking and segmentation, providing both the bounding box and the segmentation binary mask of target satellites in space. Different from on Earth, the object moved freely in space due to low gravity, and the mal-functioned satellites may spin rapidly, as shown in Figure 1(a). This adds difficulties in tracking as the template may vary significantly between frames and bounding box from tracking algorithm may not be accurate. Moreover, the target satellite appears in the image sequence may also contain scale variation and appearance deformation, as demonstrated in Figure 1(b) and (c). The size of the satellite can be drastically increase or decrease because of the largely varying chaser-target distances. If the satellite is too small, as shown in Figure 1(b), it would be hard for segmentation algorithm to identify the correct boundary. The deformation of the satellite is challenging for template update in tracking. Apart from the variation of target satellite itself, the background and illumination variation are problems to cope with. As the satellites are moving very fast in space, the background and illumination condition are changing all the

[#]These authors contributed equally to this work
*Corresponding author, email: wanxue@csu.ac.cn
*Research supported by NSFC (Natural Science Foundation of China) project 41801400 and Advancing Research of Technology and Engineering Center for Space Utilization, Chinese Academy of Science project CSU-QZKT-2018-15.

Peizhuo Li is with the University of Chinese Academy of Sciences, Technology and Engineering Center for Space Utilization, Chinese Academy of Sciences, Beijing, China and Key Laboratory of Space Utilization, Chinese Academy of Sciences, Beijing, China (Email: lipeizhuo18@mails.ucas.ac.cn)

Yunda Sun is with the University of Chinese Academy of Sciences, Technology and Engineering Center for Space Utilization, Chinese Academy of Sciences, Beijing, China and Key Laboratory of Space Utilization, Chinese Academy of Sciences, Beijing, China (Email: sunyunda18@mails.ucas.ac.cn)

Xue Wan is with the Technology and Engineering Center for Space Utilization, Chinese Academy of Sciences, Beijing, China and Key Laboratory of Space Utilization, Chinese Academy of Sciences, Beijing, China (Email: wanxue@csu.ac.cn)

time, as shown in Figure 1(d) and (e). The background and illumination variation are challenging for segmentation as the target satellites can be hard to be distinguished when backgrounds are full of clouds or in very weak illumination conditions.

As summarized in TABLE I, tracking and segmentation have their own merits and drawbacks toward the challenges of space target, background and illumination variation. To combine their advantages for robust trajectory and boundary estimation of orbiting satellite, in this paper, we proposed a Saliency-based Semi-supervised Learning for Tracking (SSLT) algorithm based on deep convolution features. To eliminate the effect of background and illumination variation, the initial bounding box is estimated by ECO[8] based tracking algorithm based on deep convolution features extracted by VGGNet[12], which proved to be robust to various environment change. Then, a saliency-based semi-supervised on-line learning approach is proposed to generate binary segmentation mask within the initial bounding box to predict accurate boundaries of the target satellite without requirement of annotated data. Finally, a fusion of tracking and segmentation is carried out to determine the final boundary of target satellite. When the target is salient, the final boundary will be refined according to the result of segmentation, whereas, the improved result will rely on the bounding box estimated by tracking algorithm.

The rest of paper is organized as follows. Section II reviews the state-of-the-art methods for on-orbit satellite tracking and segmentation. The SSLT algorithm is proposed in section III, and is compared with state-of-the-art tracking methods using an on-orbiting satellite tracking dataset in section IV. The paper concluded in section V.

## II. RELATED WORK

Visual tracking of satellite is a foundational problem in vision-based orbit servicing. It can be defined as the task that estimating trajectory of a target satellite in an image sequence. Most studies in orbiting satellite tracking are focused on low-level features, such as edges and key points. Popular algorithms for edge and key point extraction includes Harris[13], Canny[14], SIFT [15] and SURF [16]. Manolis Lourakis et al.[17] proposed an algorithm that tracks the orbiting satellites based on image edges. Low-level features, however, are not truly robust for satellite tracking especially when background and illumination condition are varying rapidly between image frames. The low-level features, edges and key points, can be largely altered if the background scenery changes from sea to desert.

Recent years, CNN-based methods have been used in computer vision, including object tracking, and turn out to have good performance. Deep convolution features, which integrates pixel-level, feature-level and semantic level features, has shown great improvement in both accuracy and robustness. TCNN[18] implements online visual tracking by managing multiple target appearance models in a tree structure. Martin Danelljan et al.[8-10, 19] apply multilayer features of CNN followed by a filter selection method has shown superior performance tested on tracking dataset. The Siamese network based trackers[20-22] formulate the visual object tracking problem as a similarity matching problem, which also achieve state-of-the-arts. However, for tracking orbiting satellite, using state-of-the-art tracking algorithms can only obtain a single tracking point or a bounding box of the object without producing a binary mask of target which can be useful to judge the status of the satellite. Moreover, according to TABLE I, the bounding box may not be very accurate when the target satellite has large appearance change, such as spin or deformation.

Driven by the increase of data and the use of deep learning, recent studies use deep convolution features, or *CNN* features, for the task of semantic segmentation and salient object detection. Mask R-CNN[23] and FusionSeg[24] are methods for object instance segmentation. Mask R-CNN can efficiently detect objects while simultaneously generating a high-quality segmentation mask for each instance. However, the model needs to be trained on a large-scale dataset. DSS[25] and OSVOS[26] use deep convolution features for salient object detection. DSS introduced short connections to the skip-layer structures within the HED[27] architecture and produced good results in salient object detection benchmarks. OSVOS tackled the task of semi-supervised video object segmentation by learning the appearance of a single annotated object of the test sequence. However, all the methods mentioned above are based on the whole image which may contain background information and is time-consuming for real-time satellite tracking.

In this paper, we proposed a SSLT algorithm relies on the initial bounding box estimated by CNN-based tracking and the segmentation binary mask generated by the saliency based semi-supervised segmentation without given annotation frame to improve the robustness and accuracy of binary masking and bounding box of orbiting target satellites. An on-line training approach is used so the SSLT algorithm is also be suitable for track and segment for other objects, as the offline training for a large-scale dataset of specific objects is not needed. Specifically, we make the following contributions:

- A saliency based semi-supervised learning is proposed for satellite segmentation without annotation dataset.
- A fusion of deep learning-based tracking and segmentation is proposed: the initial bounding box extracted by tracking is expanded and then used as input for segmentation to increase the speed and accuracy of segmentation; the final bounding box is refined based on the results of both tracking and segmentation, which is suitable for other objects not just for the satellites.

TABLE I. ADVANTAGES AND DISADVANTAGES OF TRACKING AND SEGMENTATION TO SCENE VARIATION IN SPACE

| Challenges | | Tracking | Segmentation |
|---|---|---|---|
| **Variation in space target** | *Rotation/spin* | NO, the bounding box may be inaccurate | YES |
| | *Scale change* | YES | NO, when the target is too small |
| | *deformation* | NO | YES |
| **Background change** | | YES | NO, when background is complex |
| **Illumination variation** | | YES | NO, when lighting is too weak |

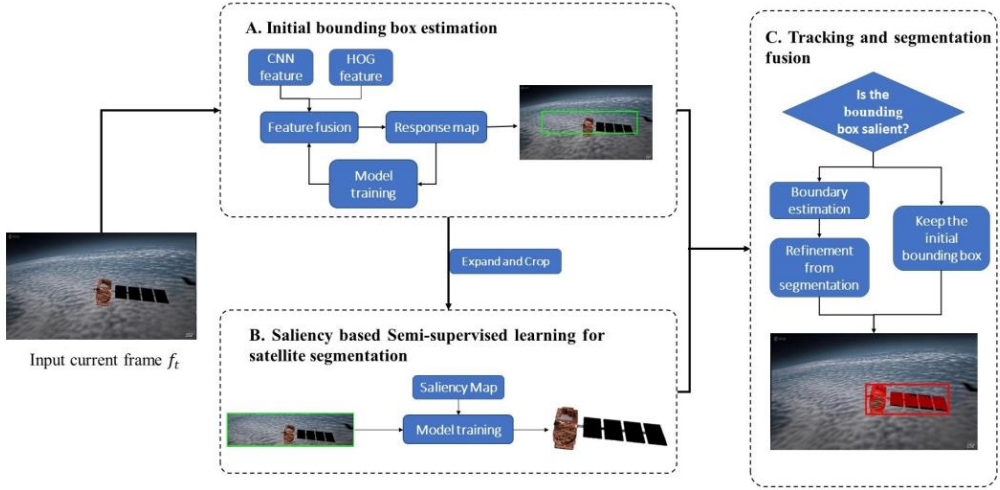

Figure 2. Pipeline of the proposed SSLT algorithm

- A new on-orbit satellite tracking and segmentation dataset is proposed including real on-orbit rendezvous and docking video from NASA, simulated satellite animation sequence from ESA and image sequences of 3D printed satellite model took in our laboratory.

## III. PROPOSED METHOD

The pipeline of the proposed SSLT algorithm is shown in Figure 2, which contains three parts: an initial bounding box estimation algorithm, a semi-supervised learning for satellite segmentation algorithm and a tracking and segmentation improvement algorithm. ECO based tracking algorithm is used to generate bounding box of satellite for segmentation. Then, the semi-supervised satellite segmentation is based on a saliency map generated from the cropped image frame determined by the initial bounding box. The final bounding box is refined by the combination of tracking and segmentation results.

### A. Initial bounding box estimation

In the proposed SSLT algorithm, the initial bounding box is estimated by ECO tracking algorithm followed by an expanding strategy and a minimal thresholding strategy. To cope with the variation in target satellite, background and illumination, CNN feature extracted by VGGNet is fused with HOG feature to train the tracking model. Here the bounding box extracted by ECO is demoted as $f_0(x, y, w, h)$, where $(x, y)$ is the upper left corner coordinate of the bounding box, $w$ and $h$ respectively are the width and the height of the initial bounding box. As mentioned in section I, the bounding box estimated from tracking is more robust to target and background variation than segmentation, however, it may lose some parts of the satellite during the tracking procedure. To solve the problem of inaccurate bounding box estimated by tracking, we designed an expanding strategy to make sure the bounding box covers the whole area of the target satellite. An expanded bounding box $f_e(x_e, y_e, w_e, h_e)$ is determined based on the central point of $f_0$ and an expanding factor $\sigma$:

$$w_e = \sigma w,$$
$$h_e = \sigma h,$$
$$x_e = x - (\sigma - 1) * w/2, \qquad (1)$$
$$y_e = y - (\sigma - 1) * h/2.$$

An optimal expanding factor $\sigma$ should be large enough to ensure the whole area of satellite are included, but small enough to exclude irrelevant target and background. In this algorithm, the expanding factor is set to ensure the frame cropped by $f_e$ is suitable for segmentation.

Generally, a large bounding box is more suitable for segmentation as segmentation is able to estimate the precise boundary within bounding box. However, if the bounding box is too small, there will be limited information for CNN-based segmentation to predict accurate boundary. Thus, based on the expanding box $f_e(x_e, y_e, w_e, h_e)$, an further bounding box $f_t(x_t, y_t, w_t, h_t)$ is determined to avoid the small bounding box for segmentation by a minimum threshold $Tr$.

$$w_t = \begin{cases} width, if\ w_e > width \\ Tr, if\ w_e < Tr \\ w_e, otherwise \end{cases},$$

$$h_t = \begin{cases} height, if\ h_e > height \\ Tr, if\ h_e < Tr \\ h_e, otherwise \end{cases}, \qquad (2)$$

$$x_t = \begin{cases} 0, if\ x_e < 0 \\ x_e, if\ x_e \geq 0 \end{cases},$$

$$y_t = \begin{cases} 0, if\ y_e < 0 \\ y_e, if\ y_e \geq 0 \end{cases}.$$

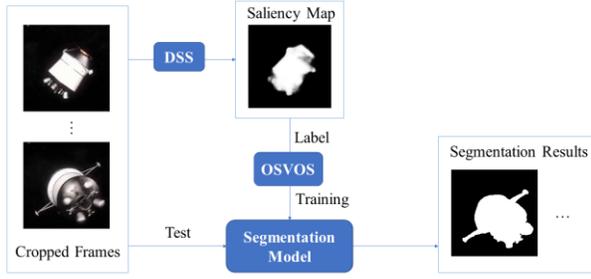

Figure 3. Pipeline of satellite segmentation

### B. Saliency based Semi-supervised learning for satellite segmentation

After the initial bounding box has been determined, a semi-supervised learning for orbiting satellite segmentation is based on OSVOS. The original OSVOS, however, is pre-trained on DAVIS 2016 training dataset, which does not include satellite scenes. To transfer the OSVOS for accurately satellite segmentation without re-training the whole network on a large annotated satellite dataset, a saliency guided semi-supervised segmentation approach is proposed, as shown in Figure 3.

Instead of using manual annotations for OSVOS segmentation model training, saliency maps produced by DSS are used as labels for on-line training. Ten frames cropped by $f_t$ are selected randomly to generate saliency maps $p_k\{k=1,2,...10\}$ using DSS in each sequence. A binary mask $saliency(i,j)$ is then generated from $p_k$ to determine the coarse boundary of segmented satellite

$$saliency(i,j) = \begin{cases} 1, if\ p(i,j) \neq 0 \\ 0, otherwise \end{cases}. \quad (3)$$

Then, the pixel number of salient area $S$ is calculated:

$$S = \sum_{i=1}^{width} \sum_{j=1}^{height} saliency(i,j). \quad (4)$$

The saliency map $p_s$ which has the largest saliency area $S$ is selected as the label for semi-supervised learning. Using the saliency map $p_s$ as the label, new weights for satellite segmentation are generated by fine-tuning the OSVOS network. For each satellite tracking scene, a customized segmentation model is generated based on the cropped frames from tracking and weights trained from saliency based semi-supervised learning. In this way, the segmentation model is flexible and versatile to cope with scene variations without annotated data.

### C. Tracking and segmentation fusion

Given the results of tracking and segmentation, our fusion method is designed as follows.

The test sequence will firstly be determined whether the frames cropped from initial ECO bounding box $f_0$ are salient or not. In this algorithm, if the bounding box is large enough for segmentation, it will be set as salient. Here we use $w$ and $h$ as the discriminant condition. If $w$ or $h$ of one frame is less than a predetermined threshold, the sequence will be determined as non-salient sequence. Otherwise, it is a salient sequence.

*Salient sequence:* In the salient sequence, the object is believed to be well segmented. The binary saliency map produced by OSVOS is used to estimate a new bounding box. Then the final boundary $f_n(x_n, y_n, w_n, h_n)$ is refined as

$$\begin{aligned} w_n &= right_i - left_i, \\ h_n &= low_j - hight_j, \\ x_n &= left_i, \\ y_n &= high_j. \end{aligned} \quad (5)$$

where the coordinates of segmented pixels in the image frame from top, bottom, left and right directions are $(high_i, high_j)$, $(low_i, low_j)$, $(left_i, left_j)$ and $(right_i, right_j)$.

*Non-salient sequence:* When the satellite is too small owing to long target-chasing distance, the segmentation of the frames cropped by expanded bounding box $f_t$ will contain many other objects or background information. In this circumstance, the result of initial ECO bounding box will have higher confidence. Consequently, we use the initial ECO bounding box $f_0$ to crop the segmentation result for refinement.

## IV. EXPERIMENT RESULTS

### A. Dataset and evaluation metrics

To evaluate the proposed SSLT method with state-of-the-art tracking and segmentation algorithms, we generated an orbiting satellite dataset of 7 sequences with 2104 image frames. Our dataset includes the on-orbit rendezvous and docking video from NASA with image size of $1280\times720$ pixels, the simulated on-orbit service video by ESA with image size of $1280 \times 720$ pixels, and a simulated SWIR (Shortwave Infrared) image sequence we took in our laboratory with image size of $640\times512$ pixels, as shown in Figure 4. A 3D satellite model was printed and then shortwave infrared images were taken by a Bobcat-640-GigE camera. Our dataset contains various challenges including variation in target, background and illumination condition as summarized in TABLE II.

TABLE II. THE CHALLENGES IN OUR TRACKING AND SEGMENTATION DATASET

| Sequence | Variation in space target | | | Background change | Illumination variation |
|---|---|---|---|---|---|
| | Rotation /spin | deformation | Scale change | | |
| 01 | ✗ | ✗ | ✓ | ✗ | ✗ |
| 02 | ✓ | ✓ | ✗ | ✗ | ✗ |
| 03 | ✓ | ✓ | ✗ | ✗ | ✗ |
| 04 | ✗ | ✗ | ✓ | ✗ | ✓ |
| 05 | ✗ | ✗ | ✓ | ✗ | ✓ |
| 06 | ✓ | ✗ | ✓ | ✓ | ✗ |
| 07 | ✓ | ✗ | ✗ | ✗ | ✗ |

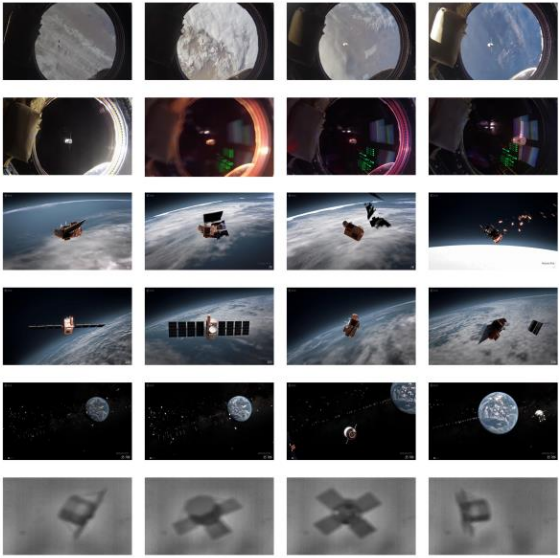

Figure 4. On-orbiting satellite tracking and segmentation dataset. *(The first two rows are real on-orbit rendezvous and docking video (credit:NASA), and the row 3 to 5 are three sequences of simulated motion and deformation of on-orbit satellites (credit:ESA), and the final row is the image sequence we took by a SWIR camera in the laborotory.)*

The evaluation is divided into two parts: the tracking performance with/without segmentation and the segmentation performance with/without tracking. To evaluate the tracking algorithms, we use the distance precision (DP) scores and the overlap precision (OP) scores. DP is defined as the percentage of frames in which the Euclidean distance between the tracker prediction and the ground truth centers is less than a threshold (100). Overlap precision (OP) is defined as the intersection-over-Uni-on (IOU) of tracker prediction and the ground truth. For segmentation performance evaluation, Structure-measure (S-measure)[28], region similarity in terms of intersection over union (J)[29], contour accuracy (F)[29] and speed are used as evaluation metrics. S-measure is a metric simultaneously evaluates region-aware and object-aware structural similarity between a saliency map (SM) and a ground truth (GT) map[28]. J and F are the measures for video object segmentation proposed in [29].

*B. Satellite tracking comparison*

In this experiment, we would like to find out whether by the combination of segmentation, the accuracy of bounding box will be improved in comparison to tracking-only algorithm. The tracking results of some challenging cases using ECO tracking and our method are shown in Figure 5. These challenging cases including two sequences which target satellite break into several parts and a quickly spinning satellite model in the laboratory. According to the tracking results in Figure 5 these large variation in target appearance will degrade the performance of ECO, as the green bounding box estimated by ECO is either too large or missing some parts of the target. In comparison, the red bounding boxes estimated by the proposed SSLT algorithm are more accurate which cover the correct boundary of the target satellite. The precision and success curve of the proposed method and original ECO, SiamFC[20] are shown in Figure 6, while the AUC (area-under-curve) scores of Precision and Success are shown in Figure 7.

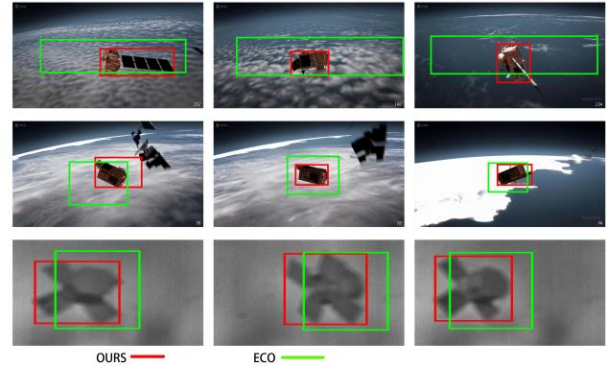

Figure 5. Our proposed method have better performance than ECO when the target have unexpected actions.(*The first and second rows shwo the satellite disintegration scene which a statellite breaks into several part. The third row shows a satellite spin scene*)

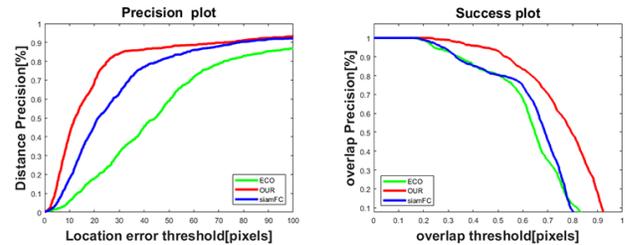

Figure 6. Precision *(Left)* and Success *(Right)* plots on the sequences. Our proposed method significantly outperforms the ECO and SiamFC on datasets

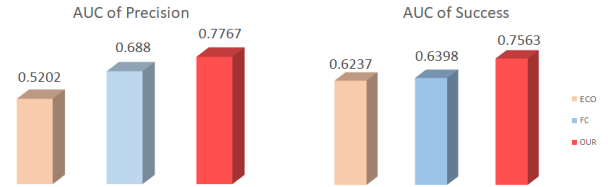

Figure 7. The comparison of AUC scores of Success and Precision

It can be concluded from Figure 6 and Figure 7 that by combining the segmentation result, the bounding box for tracking is more accurate especially when the target satellite has large appearance change such as spinning or deformation. In terms of AUC of Success, SSLT is 0.13 higher than original ECO, 0.12 higher than SiamFC. For the AUC of Precision，SSLT is 0.26 higher than original ECO, 0.09 higher than SiamFC.

*C. Satellite segmentation comparison*

Then, the proposed method is compared to state-of-the-art segmentation algorithms to find out if the performance of segmentation could be improved by initial bounding box estimated by tracking. Some of the comparison results are shown in Figure 8.

The images in the first row are ground truth, and the second and third rows are segmentation results by our algorithm and OSVOS respectively. Figure 8 shows that when the satellite is small, the original OSVOS may detect other objects as satellites, as the segmentation is based on the whole image. By utilizing the bounding box estimated by tracking, the segmentation accuracy can be largely improved compared to original OSVOS. However, our segmentation method also has limitations, for example, in the sequences of complex background and illumination variation, the solar panels of satellites may not be segmented as part of satellites by our method, because the solar panels are not salient in DSS.

TABLE III shows the average accuracy and speed of our method in comparison with OSVOS and DSS. Our method has significant improvement in both accuracy and speed compare to two state-of-the-art segmentation methods. The S-measure of our method is 0.25 higher than OSVOS and 0.31 higher than DSS. In terms of similarity $J$ and contour accuracy $F$, our method is 0.40 and 0.49 higher than OSVOS, and approximately 0.5 higher than DSS. Speed is measured in frames per second (fps). By using the result from tracking bounding box, the speed of our method achieves 4 times faster than OSVOS and 5 times faster than DSS.

It can be concluded from the experiment results in this section that by combining the tracking result into segmentation, the binary masking results of satellite segmentation can be largely improved in the aspect of accuracy and speed, especially in the case of varying background and illumination.

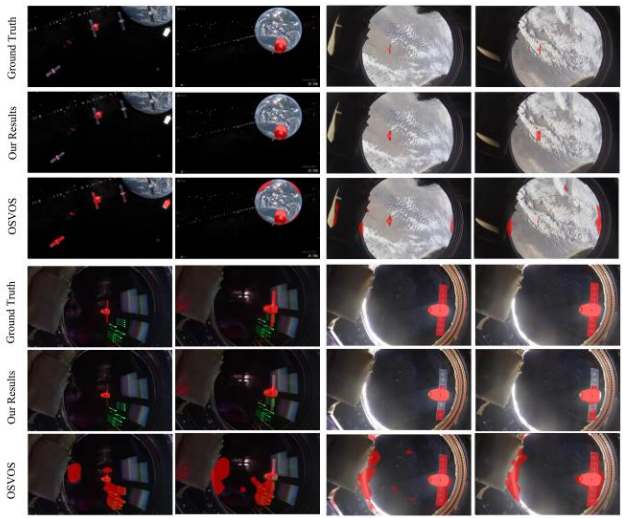

Figure 8. The results of our proposed method compared with ground truth and OSVOS in four sequences. *(The first row are ground truth. The second row are our results. The third row are OSVOS's results)*

TABLE III. THE EVALUATION OF OUR RESULTS WITH OSVOS ONLY AND DSS ONLY.

| Measures | DSS | OSVOS | OURS |
|---|---|---|---|
| S-measure↑ | 0.5553 | 0.6171 | **0.8624** |
| $J_M$↑ | 0.1552 | 0.2582 | **0.6618** |
| $J_O$↑ | 0.1279 | 0.1628 | **0.8605** |
| $J_D$↓ | -0.2103 | **-0.4428** | -0.2541 |
| $F_M$↑ | 0.2365 | 0.3437 | **0.8351** |
| $F_O$↑ | 0.0930 | 0.1628 | **0.9651** |
| $F_D$↓ | -0.3207 | **-0.3741** | -0.1588 |
| Speed | 2.588 | 3.185 | **12.605** |

## V. CONCLUSION

Visual tracking and segmentation of on-orbit satellite is vital for on-orbiting manipulation by space robots. In this paper, a SSLT algorithm has been proposed which integrates CNN-based tracking and saliency based semi-supervised on-line segmentation. Comparison experiments with state-of-the-art tracking and segmentation algorithms demonstrates the superior performance of the proposed SSLT method. Using a saliency-guided on-line learning approach, our method does not require any annotated data which can be further applied to general object tracking and segmentation. Future work will be carried out to further investigate how to recognize and locate different components in the target satellites.

ACKNOWLEDGMENT

The authors would like to thank Mingfei Han, Qingling Jia, Chenhui Wang and Huijiao Qiao for helping with image sequence taken experiment and Shiyu Xuan for valuable suggestion in satellite tracking algorithm.